\documentclass{article}

    \usepackage[preprint]{neurips_2024}

\usepackage[utf8]{inputenc} 
\usepackage[T1]{fontenc}    
\usepackage{hyperref}       
\usepackage{url}            
\usepackage{booktabs}       
\usepackage{amsfonts}       
\usepackage{nicefrac}       
\usepackage{microtype}      
\usepackage{xcolor}         

\usepackage{amsmath}
\usepackage{algorithm}
\usepackage{algorithmic}

\usepackage{multirow}
\usepackage{graphicx}
\usepackage{wrapfig}
\usepackage{amsthm}
\usepackage{subcaption}

\title{Leader Reward for POMO-Based Neural Combinatorial Optimization}

\author{%
  Chaoyang Wang \\
  Fudan University\\
  \texttt{chaoyangwang22@m.fudan.edu.cn} \\
  \And
  Pengzhi Cheng \\
  Fudan University \\
  \texttt{pzcheng23@m.fudan.edu.cn} \\
  \AND
  Jingze Li \\
  Fudan University \\
  \texttt{jingzeli20@fudan.edu.cn} \\
  \And
  Weiwei Sun \\
  Fudan University \\
  \texttt{wwsun@fudan.edu.cn} \\
}

\begin{document}

\maketitle

\begin{abstract}
  Deep neural networks based on reinforcement learning (RL) for solving combinatorial optimization (CO) problems are developing rapidly and have shown a tendency to approach or even outperform traditional solvers. However, existing methods overlook an important distinction: CO problems differ from other traditional problems in that they focus solely on the optimal solution provided by the model within a specific length of time, rather than considering the overall quality of all solutions generated by the model. In this paper, we propose Leader Reward and apply it during two different training phases of the Policy Optimization with Multiple Optima (POMO) \citep{Kwon_Choo_Kim_Yoon_Gwon_Min_2020} model to enhance the model's ability to generate optimal solutions. This approach is applicable to a variety of CO problems, such as the Traveling Salesman Problem (TSP), the Capacitated Vehicle Routing Problem (CVRP), and the Flexible Flow Shop Problem (FFSP), but also works well with other POMO-based models or inference phase's strategies. We demonstrate that Leader Reward greatly improves the quality of the optimal solutions generated by the model. Specifically, we reduce the POMO's gap to the optimum by more than 100 times on TSP100 with almost no additional computational overhead.
\end{abstract}

\section{Introduction}
\label{introduction}

Efficient methods for solving combinatorial optimization (CO) problems are widely used in industry, including in deliveries, vehicle routing, production scheduling processes, and other real-world scenarios, all of which hold significant value. However, many CO problems are NP-hard, which means they cannot be quickly solved to an optimal solution. Heuristic solvers such as LKH \citep{helsgaun2000effective,helsgaun2017extension} and HGS \citep{konstantakopoulos2022vehicle} have been proposed and are designed by experts based on domain knowledge tailored to specific problems. However, they are less efficient at solving larger-scale CO problems.

Recently, an increasing amount of work has been proposed that uses neural networks to solve CO problems like the Traveling Salesman Problem (TSP) and the Capacitated Vehicle Routing Problem (CVRP), namely Neural Combinatorial Optimization (NCO) \citep{bello2016neural}. These works mainly consist of methods based on supervised learning (SL) and reinforcement learning (RL). While SL-based methods require a large amount of labeled data, which is time-consuming for the solver to provide, RL-based methods are more attractive and promising. They not only do not rely on the expert's domain knowledge and labeled datasets, but also are capable of providing high-quality approximate solutions in a short time. 

One of the RL-based methods is the construction method, which utilizes transformer architecture \citep{DBLP:conf/nips/VaswaniSPUJGKP17} to generate solutions sequentially through a decoder. It computes the reward function and uses the REINFORCE algorithm \citep{DBLP:journals/ml/Williams92} to train the model. POMO \citep{Kwon_Choo_Kim_Yoon_Gwon_Min_2020} takes advantage of the symmetry in the CO problems. They solve the same CO problem from different perspectives and propose a new baseline suitable for the REINFORCE algorithm, which greatly reduces variance and improves the stability of training, making it one of the mainstream approaches. Several subsequent works have adopted it and, based on POMO, have optimized the performance of the inference phase or improved the model's ability to generalize across distributions and scales. Some other works apply this approach to other CO problems like the flexible flow shop problem (FFSP) and the asymmetric traveling salesman problem (ATSP).

However, such a routine does not take the special aspects of the CO problem into account. For a given CO problem, it focuses only on the optimal solution among all the solutions provided by the model after a period of inference, regardless of the quality of the other solutions. Thus, a good NCO model should value the quality of the best solution (Leader) among all the solutions given, rather than the average quality of all solutions generated. Moreover, when the model repeatedly infers the same problem, it should be able to explore more new solutions, even if the average quality of the solutions produced by each inference might decrease as a result.

In this paper, we propose Leader Reward, which changes the advantage function in the REINFORCE algorithm, and we apply it to two different training phases to motivate the model for more exploration and to place greater emphasis on the leader solution. The implementation of Leader Reward is straightforward, requiring few modifications to POMO’s advantage function and training process. By leveraging the multi-perspective problem-solving properties of POMO, it effectively integrates with POMO-based models and other inference strategies such as Simulation-Guided Beam Search (SGBS) \citep{Choo_Kwon_Kim_Jae_Hottung_Tierney_Gwon_2022} and Efficient Active Search (EAS) \citep{Hottung_Kwon_Tierney_2021}.

We demonstrate that Leader Reward provides a method to balance exploration and exploitation for POMO. Experimental results indicate that the leader solution often results from stochastic variation, suggesting it is an unbiased, correct, and under-explored direction for the model. We assess the effectiveness of Leader Reward across various CO problems, such as TSP, CVRP, and FFSP, as well as different POMO-based models like MVMoE \citep{zhou2024mvmoe} and Omni-VRP \citep{Zhou_Wu_Song_Cao_Zhang_2023}. The results show that Leader Reward significantly enhances model performance with almost no additional computational overhead. Particularly, we reduce POMO's gap to the optimum by more than 100 times on TSP100, a much greater improvement than that achieved by other studies.


\section{Related work}
\label{related_work}

\paragraph{Neural Combinatorial Optimization}
The initial attempt to solve CO problems using neural network methods was made by \citet{vinyals2015pointer}. They proposed the Pointer Network, which selects a member from the input sequence with a pointer and solved the problem of variable-size output dictionaries based on SL. As the model continuously constructs a complete solution from an initial point, it is known as the construction method. \citet{bello2016neural} suggested using RL for training because SL requires optimal solutions of CO problems as labels. Whereas CO problems are mostly NP-hard and difficult to solve optimally, it is feasible to compute the quality of a solution and design a reward function. \citet{Kool_Hoof_Welling_2018} proposed AM which modified the model to include attention layers and used the REINFORCE algorithm to train the model, employing a simple greedy rollout as a baseline through the policy gradient. \citet{Kwon_Choo_Kim_Yoon_Gwon_Min_2020} exploited the symmetry in the CO solution representation and proposed POMO which forced the model to solve the same problem from different starting points and used the average of these solutions as the baseline in the REINFORCE algorithm. This approach directly reduces the high variance of different trajectories and improves the training speed and stability of RL.

There is another type of method based on improvement which starts with a random initial solution and iteratively improves it. L2I \citep{lu2019learning} learn to refine the solution with an operator. NeuRewriter \citep{chen2019learning} learns a strategy for selecting heuristics and rewrites the current solution. Some other works improve local search or refine strategies
\citep{kim2021learning, hudson2021graph, xin2021neurolkh, Ma_Li_Cao_Song_Zhang_Chen_Tang_2021, zheng2023reinforced, ma2024learning}. However, these methods are often limited by search efficiency and high time overhead of inference.
\paragraph{POMO-Based Methods}
As POMO is a very efficient construction method that can solve hundreds of instances in a few seconds, there has since been a lot of work either exploiting the symmetry of this solution representation or extending and optimizing POMO. \citet{DBLP:conf/nips/KwonCYPPG21} applied this symmetry in the CO problem to the MatNet model, making it useful for the FFSP problem and the ATSP problem as well. Sym-NCO \citep{Kim_Park_Park_2022} exploits symmetries such as rotational invariance and reflection invariance that can greatly improve the generalization of POMO. ELG-POMO \citep{gao2023towards} devises an auxiliary strategy for learning from local transferable topological features and integrates it with typical construction policies to improve the generality of the model through joint training. Omni-VRP \citep{Zhou_Wu_Song_Cao_Zhang_2023} proposed a generic meta-learning framework to simultaneously improve the generalization of models in terms of size and distribution and applied it to POMO.

\paragraph{Inference Phase Techniques}

There are many techniques applied to the inference phase as well. When a model is trained, these inference techniques can better help the model to find the optimal solution. Active Search \citep{bello2016neural} can help models to optimize the parameters of the pointer network using RL in the test phase. SGBS \citep{Choo_Kwon_Kim_Jae_Hottung_Tierney_Gwon_2022} is able to examine candidate solutions in a fixed-width tree search. \citet{Hottung_Kwon_Tierney_2021} proposed to drastically reduce the time overhead by modifying only a fraction of the model parameters during RL training in the testing phase, known as EAS.

Our method is also based on the symmetry proposed by POMO, as we take advantage of the fact that POMO solves the same problem from different perspectives. This means that our method combines well with POMO-based methods, as well as lots of inference phase techniques, and is able to work together to improve the performance of the model.

\section{Preliminary}
\label{preliminary}

Construction-based NCO method is to train a neural network \(\pi_\theta\) with learnable weights \(\theta\). For a CO problem, the model \(\pi_\theta\) can recursively generate a solution (or a trajectory) \(\tau\) using sampling rollout or greedy rollout, denoted as \(\tau \sim \pi_\theta\).

The probability that the model makes an action \(a\) in state \(s\) is denoted as \(\pi_\theta(a|s)\), and the probability that the model generates a solution \(\tau=(a_1, a_2,\dots,a_n)\) can be computed as \(p_\theta(\tau|s) = \prod_{i=1}^n \pi_\theta(a_i|s_i) \). For each solution \(\tau\) generated by the model, the reward for that solution can be computed by the reward function \(R(\tau)\). Note that the reward is only given when the model has finished generating a solution. Given a problem instance  \(s\) randomly generated from the problem distribution \(\mathcal{D}\), we want to find the \(\theta\) that maximize the reward \({\operatorname {arg\,max}}_\theta\ \mathbb{E}_{\tau \sim \pi_\theta(s)}(R(\tau))\)

We use the REINFORCE algorithm to learn the weight parameter \(\theta\) of the model. Specifically, we need to maximize the objective function \(\mathcal{L}(\theta) = \mathbb{E}_{\tau \sim \pi_\theta(s)}(R(\tau)-b)\), where \(R(\tau)-b\) is the advantage function, and \(b\) is a baseline which is used to reduce the variance. We use the gradient ascent \(\nabla_\theta \mathcal{L}(\theta) =  (R(\tau)-b)  \nabla_\theta \log p_\theta ( \tau|s )\) to update the weight parameter \(\theta\) of the model.


\section{Methods}
\label{methods}

\subsection{Leader Reward in the main training phase}
\label{section4.1}

Inspired by POMO, we will solve the same CO problem many times from different perspectives, e.g., the TSP problem will choose different points as starting points, and the FFSP problem will choose different orders for assigning work to the machines. One of the biggest differences between CO problems and other problems is that for a given specific problem, we will only focus on the best of all the solutions, and it does not matter how good the quality of the other solutions given by the model are, except for the best one. This means that the measure of whether a model is good or bad should not be the expected quality of all the solutions given by the model \({\operatorname {arg\,max}}_\theta\ \mathbb{E}_{\tau \sim \pi_\theta(s)}(R(\tau))\), but the expected quality of the best solution after the model solves a particular CO problem multiple times \({\operatorname {arg\,max}}_\theta\ \mathbb{E}_{\tau_1,\dots,\tau_n \sim \pi_\theta(s)}\max(R(\tau_1), R(\tau_2),\dots, R(\tau_n))\). The objective function \(R(\tau) - b\) based on POMO does not allow the model to learn the importance of the leader solution.


We therefore propose the Leader Reward. Specifically, after rolling out a set of trajectories \(\{\tau_1, \tau_2, \cdots, \tau_n\}\) by sampling from the same problem \(s_i\), we compute the new advantage function using the following method:

\begin{equation}
\label{equ1}
{{\mathcal{A}_\mathrm{Leader}}}_i^j=
\begin{cases}
\alpha\times (R(\tau_i^j)-b_i)  & \text{if } j=\underset{j}{\operatorname {arg\,max}}(R(\tau^j_i)-b_i)\\
R(\tau^j_i)-b_i  & \text{if } j\neq \underset{j}{\operatorname {arg\,max}}(R(\tau^j_i)-b_i).
\end{cases}
\end{equation}

In Equation~\ref{equ1}, \(\alpha>1\), which is a hyperparameter that controls the extra reward earned by the leader. 
We will take the average reward of all the trajectories sampled from the same problem as the baseline \(b_i\), and then add an extra Leader Reward to the leader trajectory. To maximize the objective function \(\mathcal{L}(\theta)\), we use a gradient ascent as follows:

\begin{equation}
\nabla_\theta \mathcal{L}(\theta) \leftarrow \frac{1}{B N} \sum_{i=1}^B \sum_{j=1}^N{\mathcal{A}_\mathrm{Leader}}_i^j \nabla_\theta \log p_\theta ( \tau_i^j |s_i ),
\end{equation}

where \( \log  p_\theta( \tau_i^j |s_i )\) represents the log probability of sampling the trajectory \(\tau_i^j\), \(B\) is the batch size, and \(N\) is the number of different trajectories sampling from the same problem. Algorithm~\ref{algo1} describes the pseudocode for applying Leader Reward to the training. It is worth noting that in practice, we divide the advantage value for all trajectories by \(\alpha\) (line 7). Since the Adam optimizer is used, multiplying the advantage function by a constant value does not change the result. However, when \(\alpha\)  is large, there are some advantages to this approach, which we will explain in Section~\ref{section4.2}.

\begin{algorithm}
\caption{Applying Leader Reward in the main training phase}
\label{algo1}
\begin{algorithmic}[1] 
\REQUIRE model parameter \(\theta\), batch size  \(B\), problem distribution  \(\mathcal{D}\), number of starting nodes  \(N\), number of training steps \(T\), Leader Reward multiplier \(\alpha\)

\FOR{\(step = 1\) to \(T\)}
    \FOR{\(i = 1\) to \(B\)}
        \STATE \(s_i \leftarrow \textbf{SampleInstance}(\mathcal{D})\)
        \STATE \(\{a_i^1, a_i^2, \dots, a_i^N\} \leftarrow \textbf{SelectStartNodes}(s_i)\)
        \STATE \(\tau_i^j \leftarrow \textbf{SampleRollout}(a_i^j, s_i, \pi_\theta) \quad \forall j \in \{1,\dots ,N\}\)
        \STATE \(b_i \leftarrow \frac{1}{N} \sum_{j=1}^N R(\tau_i^j)\)
        \STATE \({\mathcal{A}_\mathrm{Leader}}_i^j \leftarrow \frac{1}{\alpha} (R(\tau^j_i)-b_i) \quad \forall j \in \{1,\dots ,N\}\)
        \STATE \(l_i^* \leftarrow \operatorname {arg\,max}_j(R(\tau^j_i)-b_i) \quad \forall j \in \{1,\dots ,N\}\)
        \STATE \({\mathcal{A}_\mathrm{Leader}}_i^{l_i^*} \leftarrow (R(\tau^{l_i^*}_i)-b_i)\)
    \ENDFOR
    \STATE \(\nabla_\theta \mathcal{L}(\theta) \leftarrow \frac{1}{B N} \sum_{i=1}^B \sum_{j=1}^N{\mathcal{A}_\mathrm{Leader}}_i^j \nabla_\theta \log p_\theta ( \tau_i^j|s_i )\  \forall i \in \{1,\dots ,B\}, \forall j \in \{1,\dots ,N\}\)
    \STATE \(\theta \leftarrow \textbf{Adam} (\theta, \nabla_\theta \mathcal{L}(\theta))\)
\ENDFOR
\ENSURE trained model parameter \(\theta\)
\end{algorithmic}
\end{algorithm}

\begin{wrapfigure}{r}{0.5\textwidth}
    \centering
    \includegraphics[width=1\linewidth]{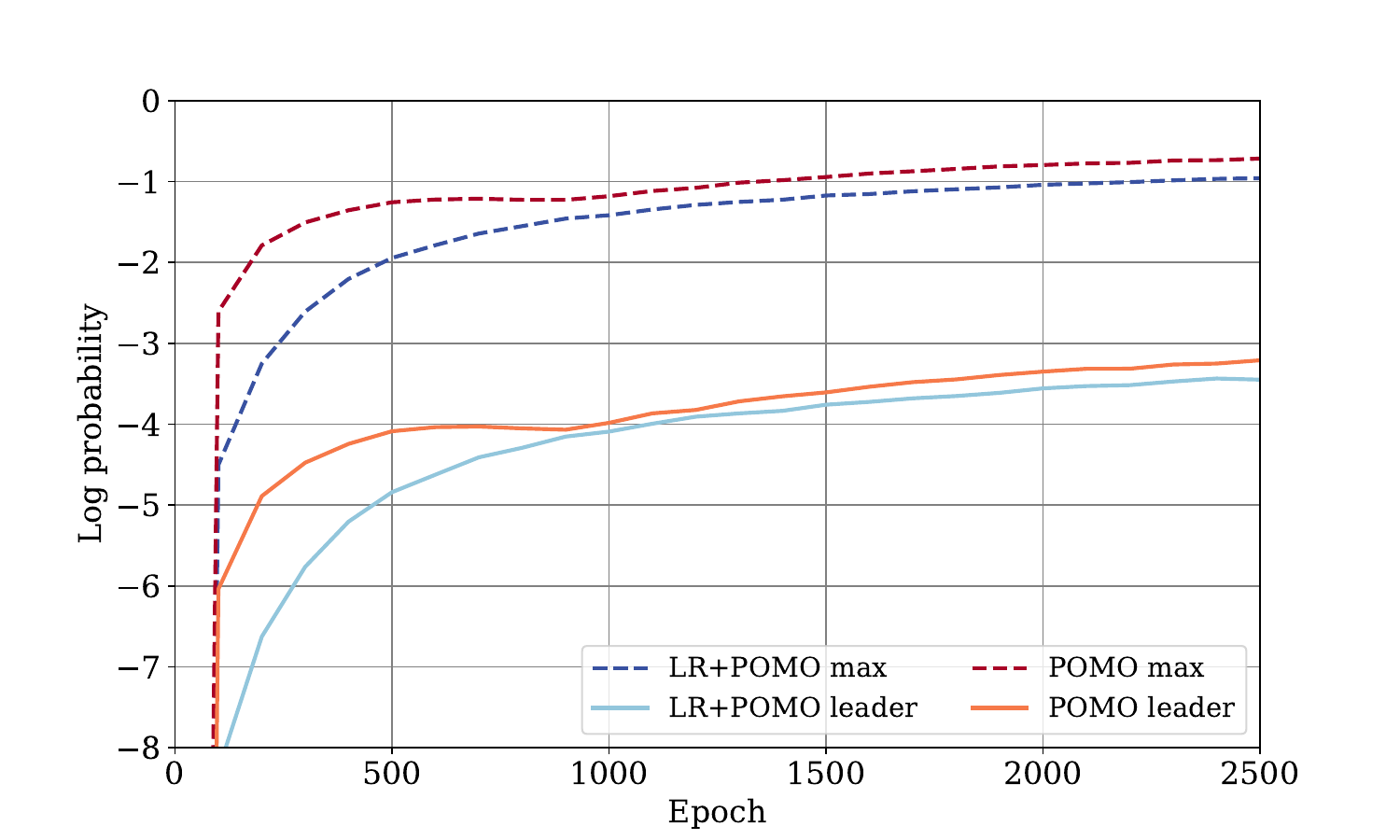}
    \caption{Log probability curve during the training phase.}
    \label{fig2}
\end{wrapfigure}

In Figure~\ref{fig2}, we show the log probability of generating the leader trajectory and the maximum log probability of 100 trajectories for each TSP100 problem during the training phase. The log probability of the leader trajectory is much smaller than the maximum log probability, implying that it is a low-probability event generated by random sampling and that the model has not yet sufficiently explored it. We also show the results after adding the Leader Reward (LR) during training, which remain the same. This is because all the datasets are randomly generated during the training phase, so no data is reused, thus reducing the risk of overfitting.

\clearpage
\paragraph{Proposition 1.}
\textit{Consider one of the steps during the generation of the solution for the model. Let \(s\) be the state given by the environment at this step, and \(a_i\) be each action the model can take under the state. Let \(\pi_\theta\) be the neural network whose parameters are \(\theta\) and learning rate is \(\gamma\). The probability of each action model can take \(\pi_\theta(a_i | s)\) is calculated from the softmax function \(\sigma(z_i)={e^{z_i}}/({\sum_j e^{z_j}})\) where \(z_i\) represents the score for each decision calculated by the model. The entropy \(H(p)\) of a probability distribution \(p\) over \(n\) actions is given by \(H(p)=-\sum_{i=1}^n p_i \log(p_i)\). Assume \(a^*\) is the learder action and \(\pi_\theta(a^* | s) < -\frac{1}{n}\sum_i\pi_\theta(a_i | s)\), and the new entropy is \(H(p')\) after giving the leader an extra reward \(r\). Then \(H(p') > H(p))\).}

See Appendix~\ref{app_proof} for the proof. Therefore, Leader Reward is also a way to increase exploration and control the model's balance between exploration and exploitation by adjusting the size of the value of \(\alpha\). In RL, exploration is crucial for effectively learning how to navigate an environment and make optimal decisions. It can help the model discover new strategies as well as handle sparse rewards. Some traditional methods, such as the Epsilon-Greedy Algorithm \citep{sutton2018reinforcement} will use a hyperparameter \(\epsilon\) to make the model explore randomly with a certain small probability, but the directions explored in this randomized exploration may be sometimes good, sometimes bad. However, the Leader Reward will guide the model in the right direction, as the leader trajectory derived from multiple rollouts of the same problem is a solution free of fluctuations and bias, as well as a new direction not fully explored by the model. This means that the quality of the model's exploration will be higher than traditional methods.

\subsection{Leader Reward in the final training phase}
\label{section4.2}

The distinction in CO problems, which focuses solely on the quality of the optimal solution, not only alters the metrics for evaluating model performance but also modifies the strategies employed by models during inference. When assessing a model's inferential capabilities, we usually consider the model's performance across various inference time durations and tailor our method to meet the demands of real-world scenarios. If enough time is available for inference, allowing the model to generate numerous trajectories through sampling rollout, a model that explores diverse potential paths and generates varied solutions will outperform others. Conversely, if the model is afforded only a limited number of attempts or even just a single try at solving a problem due to time constraints, a more robust model is probabilistically favored to provide a reasonably good approximate solution.

\begin{algorithm}
\caption{Applying Leader Reward in the final training phase}
\label{algo2}
\begin{algorithmic}[1] 
\REQUIRE learning rate  \(\gamma\), model parameter \(\theta\), batch size  \(B\), problem distribution  \(\mathcal{D}\), number of starting nodes  \(N\), number of training steps \(T'\), Leader Reward multiplier \(\alpha\)
\FOR{\(step = 1\) to \(T'\)}
    \FOR{\(i = 1\) to \(B\)}
        \STATE \(s_i \leftarrow \textbf{SampleInstance}(\mathcal{D})\)
        \STATE \(\{a_i^1, a_i^2, \dots, a_i^N\} \leftarrow \textbf{SelectStartNodes}(s_i)\)
        \STATE \(\tau_i^j \leftarrow \textbf{SampleRollout}(a_i^j, s_i, \pi_\theta) \quad \forall j \in \{1,\dots ,N\}\)
        \STATE \(b_i \leftarrow \frac{1}{N} \sum_{j=1}^N R(\tau_i^j)\)
        \STATE \(l_i^* \leftarrow \operatorname {arg\,max}_j(R(\tau^j_i)-b_i) \quad \forall j \in \{1,\dots ,N\}\)
        \STATE \({\mathcal{A}_\mathrm{Leader}}_i^{l_i^*} \leftarrow (R(\tau^{l_i^*}_i)-b_i)\)
    \ENDFOR
    \STATE \(\nabla_\theta \mathcal{L}(\theta) \leftarrow \frac{1}{B N} \sum_{i=1}^B \sum_{j=1}^N{\mathcal{A}_\mathrm{Leader}}_i^j \nabla_\theta \log p_\theta ( \tau_i^j|s_i )\  \forall i \in \{1,\dots ,B\}, \forall j \in \{1,\dots ,N\}\)
    \STATE \(\theta \leftarrow \textbf{Adam} (\theta, \nabla_\theta \mathcal{L}(\theta),\gamma)\)
\ENDFOR
\ENSURE trained model parameter \(\theta\)
\end{algorithmic}
\end{algorithm}

As mentioned in Section~\ref{section4.1}, applying different values of \(\alpha\) during training alters the extent to which the model explores, leading to a final model that tends towards being either aggressive or conservative. This approach can also be employed in the final training phase. Hence, we propose Algorithm~\ref{algo2}, a method based on Leader Reward, to adjust the conservativeness of the model. This algorithm could enhance the model's aggressiveness by setting \(\alpha = +\infty\) and balance the specific degrees of aggression and conservatism by adjusting the learning rate \(\gamma\), ensuring the model converges to an appropriate state. It just requires a small amount of training time, converging in just 100 epochs on TSP100, which is approximately one-thirtieth of the entire training process.

\begin{wrapfigure}{r}{0.5\textwidth}
    \centering
    \includegraphics[width=1\linewidth]{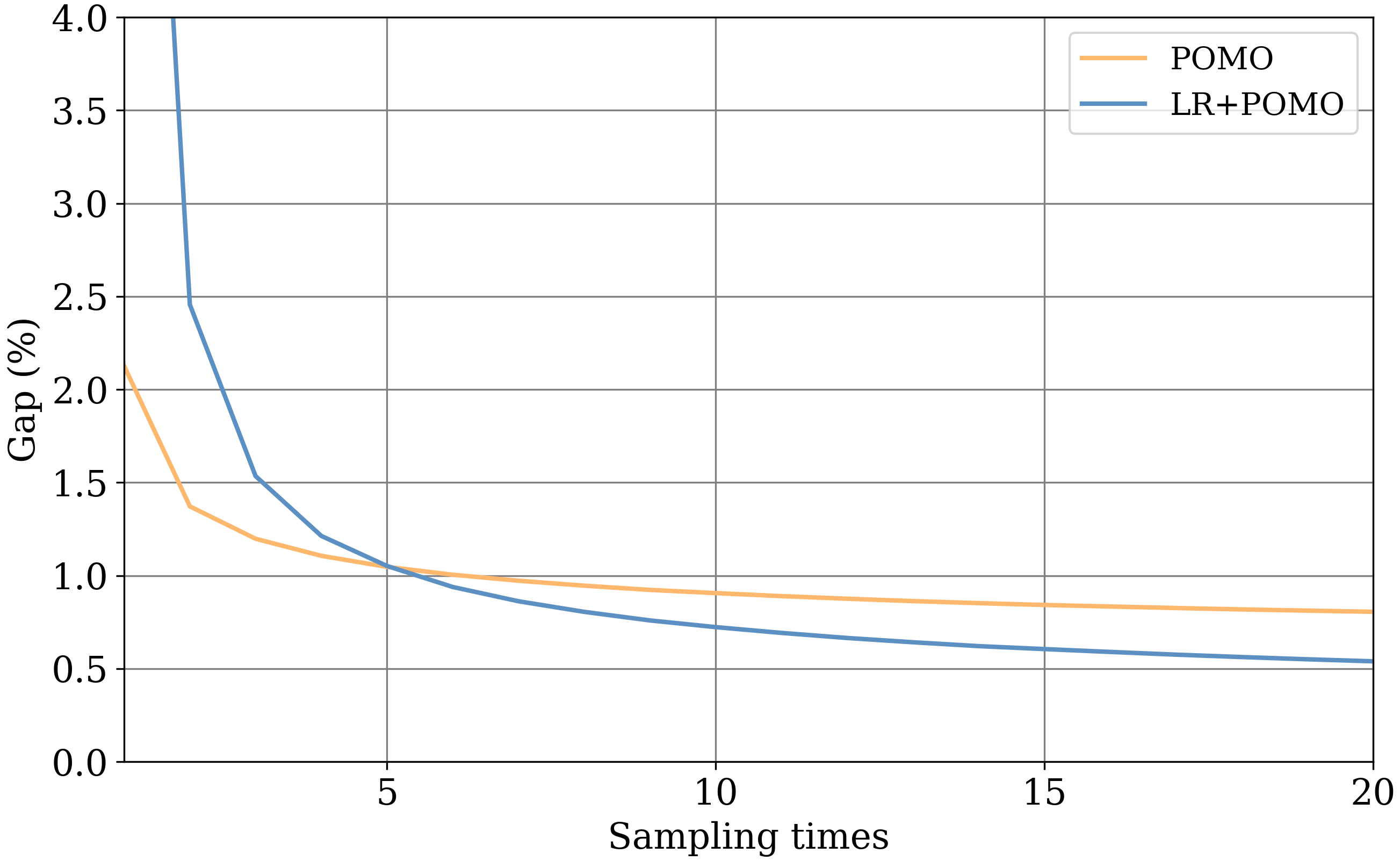}
    \caption{Gap of different sampling times on 10,000 instances of TSP100. }
    \label{fig3}
\end{wrapfigure}

Figure~\ref{fig3} shows the gap between the solutions generated by the model and the optimal solution when applying Algorithm~\ref{algo2} to solve the same problem by sampling multiple times. It is observed that this algorithm results in a decline in the average quality of solutions produced by the model, as the gap of the solution provided by the model escalates from 2.13\% to 8.52\% when solving the problem just once. This escalation occurs because, when a model solves the same problem only a limited number of times, the emphasis is on the average quality of solutions it provides. Conversely, as the model solves the same problem more frequently, a more aggressive model becomes advantageous.

Given that POMO leverages the symmetry of solution representations and data augmentation, POMO-based methods typically involve solving the same problem a large number of times during the inference phase. Taking the TSP100 as an example, these methods will sample the same problem 800 times. This algorithm is designed to leverage POMO’s capability of repeated problem-solving, enhancing the model's aggressiveness at the expense of deteriorating the average solution quality. The trade-off aims to improve the probability of generating high-quality, optimal solutions.

\subsection{Leader Reward + SGBS + EAS}

As POMO is a highly efficient construction method, recently numerous improvements have been made to enhance its performance. One such improvement is SGBS + EAS, an optimization strategy tailored for the POMO's inference phase. Our method can also be applied to the inference phase and works well with SGBS + EAS, further reducing the gap between the model's solution and the optimum.


SGBS is an inference phase technique that enhances the quality of solutions generated by the POMO model. During the solution generation process,  it produces \(\beta \times (\gamma-1)\) directions at each step, and retains the best \(\beta\) among them through a greedy rollout mechanism. This method aims to generate the best \(\beta\) solutions at the same time, thereby improving the quality of the solutions generated by the model. Such an algorithm, capable of efficiently producing multiple solutions, enables us to attain a superior solution within a short inference time while also emphasizing the exploratory performance of the model. In contrast to the previous practice of pursuing a single greedy rollout to obtain a nearly approximate solution, we now, with the efficient inference strategy of SGBS, prefer the model to capitalize on the greater opportunities provided by SGBS to fully explore the potential of other solutions.




EAS is a method that trains on a test set by using POMO's pre-trained model and updates the parameters during testing to enhance performance. It leverages the fact that the problem only requires an optimal solution and, given sufficient inference time, allows for tuning the model during this period. EAS accelerates the training process by updating only a subset of parameters. Since EAS involves training on a test set, it can be integrated with the approach described in Section~\ref{section4.1}. This integration treats EAS as the main training phase and uses Leader Rewards to guide the model in a correct new direction during training.

SGBS and EAS collaborate in a way that SGBS assists EAS by sharing incumbent solutions. SGBS helps EAS escape local optima by discovering superior solutions through efficient generative methods that leverage new paths explored by EAS during training. This method alternates between EAS and SGBS in sequence. However, SGBS has a significant time overhead, as one iteration of SGBS(4,4) on CVRP100 takes about 6 times longer than one iteration of EAS, and one iteration of SGBS(10,10) on TSP100 takes about 20 times longer than one iteration of EAS, which makes this strategy less viable when training time is limited.  Experimentally, we found that increasing the number of iterations of EAS improves model performance when time is limited, but at the expense of making the model more likely to fall into a local optimum and perform worse in the long term. However, since EAS operates in the testing phase, where time is precious, we believe that sacrificing a little long-term performance for a reduction in short-term performance is worthwhile.


There are also lots of other works based on POMO, such as MatNet, Omni-VRP, and MVMoE. Our approach is an optimization of the POMO training approach, and we find that it simultaneously benefits these POMO-based models. The experimental results are shown in Section~\ref{experiments}. This demonstrates the generalizability of our approach.

\section{Experiments}
\label{experiments}

The training and testing were conducted on a single Nvidia RTX 4090 GPU. To evaluate the effectiveness of Leader Reward (abbreviated as LR in the table), we modified POMO (MIT license) \citep{Kwon_Choo_Kim_Yoon_Gwon_Min_2020} and assessed its performance on the TSP and CVRP problems. We also applied our method to MatNet (MIT license) \citep{DBLP:conf/nips/KwonCYPPG21} and tested its performance on the FFSP problem to show our method's applicability to models that utilize the symmetric representation of solutions in POMO. Since the SGBS+EAS optimization is relevant to all three problems, we also evaluated the performance after implementing SGBS+EAS (MIT license) \citep{Choo_Kwon_Kim_Jae_Hottung_Tierney_Gwon_2022} in the inference phase. To further demonstrate the applicability of our method to other POMO-based models, we tested Leader Reward on two additional models, MVMoE (MIT license) \citep{zhou2024mvmoe} and Omni-VRP (MIT license) \citep{Zhou_Wu_Song_Cao_Zhang_2023}. The results can be found in Appendix~\ref{mvmoeomni}. We also tested the performance of Leader Reward on the realistic datasets TSPLib \citep{reinelt1991tsplib} and CVRPLib \citep{uchoa2017new}; details of these tests can be found in Appendix~\ref{tsplib}. As some of the results in the table were obtained from different hardware (v3-8 TPU or Nvidia A100 GPUs), we've marked their time consumption with \(*\) for a fair comparison.

\paragraph{TSP}

The TSP problem can be considered one of the benchmark problems among CO problems, and many NCO models are tested for performance on the TSP problem. In the TSP, there are multiple points in a two-dimensional space, and the distance between any two points is measured using Euclidean distance. The goal is to find the minimal distance required to start from any one point, pass through each point exactly once, and return to the starting point. To maintain consistency with previous experiments, we used the test set generated by \citet{Choo_Kwon_Kim_Jae_Hottung_Tierney_Gwon_2022}. Specifically, 10,000 instances of TSP100 were generated using the random seed 1234, and 1,000 instances each of TSP150 and TSP200 were generated using the random seed 1235.

We use the results of exact-solver Concorde \citep{cook2011traveling} as a standard for comparison, and we also give the results of LKH3. For the NCO method, we give results for DPDP \citep{DBLP:conf/cpaior/KoolHGW22}, COMPASS \citep{DBLP:conf/nips/ChalumeauSBGPLB23}, Poppy \citep{DBLP:conf/nips/GrinsztajnFSBB23}, and POMO. However, since our optimization is based on POMO, our main emphasis is on the magnitude of improvement in POMO. Just as POMO can be optimized using SGBS+EAS, we also present the results of combining Leader Reward with SGBS+EAS, as well as the results for different lengths of inference time.

As POMO trained a TSP100 model for a total of 3050 epochs, we also kept the overall training time consistent to ensure a fair comparison. Specifically, we use \(\alpha = 40\) when applying Leader Reward in the main training phase and trained POMO for 2900 epochs, in the final training phase we set \(\alpha = +\infty\) trained for 150 epochs. In the inference phase, we combined Leader Reward with SGBS+EAS. We selected the hyperparameters \(\beta = 10\) and \(\gamma = 10\) for SGBS. For EAS, we set the parameter of Leader Reward to \(\alpha=40\) and performed SGBS after every 20 EAS iterations.

Table~\ref{tsp-table} shows the performance of Leader Reward on the TSP problem. The result for LR+SGBS is not included in the table because we found that LR+SGBS+EAS performs better than LR+SGBS in terms of both time consumption and the gap. It can be found that Leader Reward can significantly improve the performance and generalization of the POMO model on the TSP. Specifically, Leader Reward reduces the POMO’s gap to the optimum by 10 times, and can be boosted up to 200 times when optimized with SGBS+EAS inference strategy. For generalization capabilities, Leader Reward can also reduce the gap by 97\% and 84\% on TSP150 and TSP200, respectively, which represents a huge improvement greater than that achieved by other studies.


\begin{table}
\caption{Experiment results on TSP}
\label{tsp-table}
\centering
\resizebox{\textwidth}{!}{
\begin{tabular}{l||clc|clc|clc}
\toprule
\multirow{3}{*}{Method}    & \multicolumn{3}{c|}{Test (10K instances)}  & \multicolumn{6}{c}{Generalization (1K instances)}         \\
& \multicolumn{3}{c|}{TSP100}                & \multicolumn{3}{c|}{TSP150}               & \multicolumn{3}{c}{TSP200}                \\
& Cost           & \multicolumn{1}{c}{Gap}               & Time & Cost           & \multicolumn{1}{c}{Gap}              & Time & Cost            & \multicolumn{1}{c}{Gap}              & Time \\
\midrule
Concorde           & 7.765          & \multicolumn{1}{c}{-}                 & 82m  & 9.346          & \multicolumn{1}{c}{-}                & 17m  & 10.687          & \multicolumn{1}{c}{-}                & 31m  \\
LKH3               & 7.765          & 0.000\%           & 8h   & 9.346          & 0.000\%          & 99m  & 10.687          & 0.000\%          & 3h   \\
\midrule
Poppy 16           & 7.770          & 0.07\%            & 1m*   & 9.372          & 0.27\%           & 20s*  &                 &                  &      \\
POMO               & 7.776          & 0.144\%           & 1m   & 9.397          & 0.544\%          & 14s  & 10.843          & 1.459\%          & 31s  \\
\textbf{LR+POMO}                 & \textbf{7.766} & \textbf{0.014\%}  & 1m   & \textbf{9.364} & \textbf{0.193\%} & 14s  & \textbf{10.792} & \textbf{0.985\%} & 31s  \\
\midrule
POMO(sampling)     & 7.771          & 0.078\%           & 3h*   & 9.378          & 0.355\%          & 1h*   & 10.838          & 1.417\%          & 3h*  \\
DPDP               & 7.765          & 0.004\%           & 2h*   & 9.434          & 0.937\%          & 44m*  & 11.154          & 4.370\%          & 74m*  \\
COMPASS            & 7.765          & 0.002\%           & 2h*   & 9.350          & 0.043\%          & 32m*  & 10.723          & 0.337\%          & 70m*  \\
SGBS+EAS(short)    & 7.770          & 0.063\%           & 37m  & 9.368          & 0.236\%          & 11m  & 10.764          & 0.718\%          & 29m  \\
SGBS+EAS           & 7.768          & 0.045\%           & 2h   & 9.359          & 0.142\%          & 2h   & 10.739          & 0.484\%          & 2h   \\
\textbf{LR+SGBS+EAS(short)} & 7.765          & 0.0007\%           & 25m  & 9.347          & 0.009\%          & 17m  & 10.701          & 0.133\%          & 37m  \\
\textbf{LR+SGBS+EAS}        & \textbf{7.765} & \textbf{0.0002\%} & 2h   & \textbf{9.346} & \textbf{0.005\%} & 2h   & \textbf{10.695} & \textbf{0.075\%} & 2h  \\
\bottomrule
\end{tabular}
}
\end{table}


\paragraph{CVRP}

In the CVRP, multiple points are distributed across a two-dimensional space, each associated with a demand value. A vehicle may depart from the depot multiple times, pass through some of these points, and then return to the depot. The total demand of the points visited in each trip must not exceed the vehicle's capacity. The goal is to find the shortest route that ensures each point is visited at least once.

We used the same test set setup as in the TSP problem, and we used the results of the heuristic solver HGS \citep{DBLP:journals/ior/VidalCGLR12, DBLP:journals/cor/Vidal22} as the baseline for comparison. For the training process, we used \(\alpha = 10\) when applying Leader Reward in the main training phase and trained the POMO model for 28,500 epochs, and in the final training phase, we set \(\alpha = +\infty\). In the inference phase, we selected the hyperparameters \(\beta=4\) and \(\gamma=4\) for SGBS. For EAS, we set the parameter of Leader Reward to \(\alpha=10\) and performed SGBS after every 3 EAS iterations.

Table~\ref{cvrp-table} displays the performance of Leader Reward on the CVRP problem. It is evident that Leader Reward enhances both the performance and generalization of the POMO model on this problem, reducing the gap by 35\%, 27\%, and 40\% for CVRP100, CVRP150, and CVRP200, respectively.


\paragraph{FFSP}

To demonstrate the applicability of Leader Reward to models that leverage the symmetry of the solution representation, we also applied it to MatNet and tested the performance on the FFSP problem.

The FFSP problem is modeled on the production scheduling process in a real manufacturing application, where each job must be performed sequentially across \(S\) stages. Each stage has \(M\) types of machines, and each job takes a different amount of time on each machine. Within the same stage, it is only necessary to work on any one of the \(M\) types of machines. The goal of the problem is to find the shortest possible time required to complete all jobs.

To maintain consistency with previous experiments, we used \(S=3\) and \( M=4\), and the possible time required for each job was a random integer from 2 to 9. We used the test set generated by \citet{DBLP:conf/nips/KwonCYPPG21} and tested how the model performed with \(N=\{20,50,100\}\) jobs. To ensure that the total training time is consistent, we chose \(\alpha=\{4, 4, 2\}\) and trained MatNet for \(\{50, 100, 150\}\) epochs. In the inference phase, we selected the hyperparameters \(\beta=5\) and \(\gamma=6\) for SGBS, and we performed SGBS after every 3 EAS iterations. We also give results for mixed-integer programming models CPLEX, and other heuristics methods.

Results in Table~\ref{ffsp-table} show that the Leader Reward method outperforms both heuristic methods and NCO methods in solving the FFSP problem and significantly improves the performance of MatNet, demonstrating the effectiveness of our method for MatNet.

\begin{table}
\caption{Experiment results on CVRP}
\label{cvrp-table}
\centering
\resizebox{\textwidth}{!}{
\begin{tabular}{l||lcc|lcc|ccc}
\toprule
\multirow{3}{*}{Method}    & \multicolumn{3}{c|}{Test (10K instances)}  & \multicolumn{6}{c}{Generalization (1K instances)}                                     \\
& \multicolumn{3}{c|}{CVRP100}               & \multicolumn{3}{c|}{CVRP150}               & \multicolumn{3}{c}{CVRP200}               \\
& \multicolumn{1}{c}{Cost}            & Gap              & Time & \multicolumn{1}{c}{Cost}            & Gap              & Time & Cost            & Gap              & Time \\
\midrule
HGS                & 15.563          & -                & 24h  & 19.052          & -                & 5h   & 21.755          & -                & 9h   \\
LKH3               & 15.646          & 0.532\%          & 6d   & 19.222          & 0.891\%          & 20h  & 22.003          & 1.138\%          & 25h  \\
\midrule
Poppy 32           & 15.73           & 1.072\%          & 5m*   & \textbf{19.50}  & \textbf{2.350\%} & 1m*   &                 &                  &      \\
POMO               & 15.754          & 1.228\%          & 1m   & 19.686          & 3.324\%          & 16s   & 23.057          & 5.983\%          & 34s   \\
\textbf{LR+POMO}                 & \textbf{15.729} & \textbf{1.064\%} & 1m   & 19.682          & 3.306\%          & 16s   & \textbf{23.012} & \textbf{5.775\%} & 34s   \\
\midrule
POMO(sampling)     & 15.663          & 0.641\%          & 6h*   & 19.478          & 2.235\%          & 2h*   & 23.176          & 6.530\%          & 5h*   \\
DPDP               & 15.627          & 0.410\%          & 23h*  & 19.312          & 1.363\%          & 5h*   & 22.263          & 2.333\%          & 9h*   \\
COMPASS(no aug)    & 15.594          & 0.198\%          & 4h*   & 19.313          & 1.369\%          & 2h*   & 22.462          & 3.248\%          & 2h* \\
SGBS+EAS(short)    & 15.605          & 0.271\%          & 2h   & 19.227          & 0.920\%          & 1h   & 22.274          & 2.382\%          & 3h   \\
SGBS+EAS           & 15.587          & 0.152\%          & 10h  & 19.154          & 0.532\%          & 3h   & 22.109          & 1.626\%          & 7h   \\
\textbf{LR+SGBS+EAS(short)} & 15.588          & 0.158\%          & 2h   & 19.160          & 0.566\%          & 1h   & 22.106          & 1.611\%          & 3h   \\
\textbf{LR+SGBS+EAS}        & \textbf{15.579} & \textbf{0.099\%} & 10h  & \textbf{19.126} & \textbf{0.389\%} & 3h   & \textbf{21.966} & \textbf{0.967\%} & 7h  \\
\bottomrule
\end{tabular}
}
\end{table}


\begin{table}
\caption{Experiment results on FFSP (1K instances)}
\label{ffsp-table}
\centering
\resizebox{\textwidth}{!}{
\begin{tabular}{l||lrc|lrc|lrc}
\toprule
 & \multicolumn{3}{c|}{FFSP20} & \multicolumn{3}{c|}{FFSP50} & \multicolumn{3}{c}{FFSP100}\\
 & \multicolumn{1}{c}{Cost} & \multicolumn{1}{c}{Gap} & Time & \multicolumn{1}{c}{Cost} & \multicolumn{1}{c}{Gap} & Time & \multicolumn{1}{c}{Cost} & \multicolumn{1}{c}{Gap} & Time \\
\midrule
CPLEX(60s) & 46.37 & 91.857\% & 17h &  \multicolumn{1}{c}{\(\times\)}&  &  & \multicolumn{1}{c}{\(\times\)} &  &  \\
CPLEX(600s) & 36.56 & 51.268\% & 167h &  &  &  &  &  &  \\
\midrule
Genetic Algorithm & 30.57 & 26.484\% & 56h & 56.37 & 16.851\% & 128h & 98.69 & 12.036\% & 232h  \\
Particle Swarm Opt. & 29.07 & 20.278\% & 104h & 55.11 & 14.239\% & 208h & 97.32 & 10.480\% & 384h  \\
\midrule
MatNet & 25.392 & 5.060\% & 2m & 49.600 & 2.817\% & 5m & 89.745 & 1.881\% & 13m  \\
\textbf{LR+MatNet} & \textbf{25.232} & \textbf{4.398\%} & 2m & \textbf{49.363} & \textbf{2.326\%} & 5m & \textbf{89.207} & \textbf{1.270\%} & 13m   \\
\midrule
MatNet(sampling) & 24.60 & 1.783\% & 10h* & 48.78 & 1.117\% & 20h* & 88.95 & 0.979\% & 40h*  \\
SGBS+EAS(short) & 24.467 & 1.233\% & 3h & 48.837 & 1.235\% & 6h & 88.980 & 1.013\% & 10h   \\
SGBS+EAS & 24.250 & 0.335\% & 15h & 48.519 & 0.576\% & 30h & 88.568 & 0.545\% & 60h   \\
\textbf{LR+SGBS+EAS(short)} & 24.395 & 0.935\% & 3h & 48.596 & 0.736\% & 6h & 88.455 & 0.417\% & 10h   \\
\textbf{LR+SGBS+EAS} & \textbf{24.169} & \multicolumn{1}{c}{\textbf{-}} & 15h & \textbf{48.241} & \multicolumn{1}{c}{\textbf{-}} & 30h & \textbf{88.088} & \multicolumn{1}{c}{\textbf{-}} & 60h  \\
\bottomrule
\end{tabular}
}
\end{table}

\section{Conclusion and Discussion}
\label{conclusion}

In this paper, we propose a new advantage function, Leader Reward, applicable to the POMO model and other models that leverage the symmetry of this solution representation. We analyze the specificity of the CO problem and apply Leader Reward during two different phases of training to improve models' performance. Experiments show that this method performs well on various CO problems (TSP, CVRP, FFSP) and with different POMO-based models (MatNet, Omni-VPR, MVMoE). It is capable of further improving the model's performance using inference strategies. Moreover, this performance enhancement incurs almost no additional computational overhead.

For potential societal impacts, we believe that the CO problem is closely related to practical challenges such as deliveries, vehicle routing, and production scheduling. With the development of NCO, models that provide solutions with greater accuracy and in less time can significantly enhance the efficiency of societal and productive activities. Therefore, we believe that the societal impacts of our work are mainly positive, with generally no negative societal impacts. In future work, we will attempt to dynamically adjust the parameter \(\alpha\) in the Leader Reward, aiming to better integrate it with other models and inference strategies.

Our code for the experiments can be found in the supplementary material.

\clearpage
\bibliography{neurips_2024}


\clearpage
\appendix

\section{Proof}
\label{app_proof}

\begin{proof}

Assume there are \(n\) actions \(\{z_1, z_2,\dots,z_n\}\) and the leader action is \(z_1\). The probability of each actions chosen by model \(\displaystyle p_i=\sigma(z_i)= \frac{e^{z_i}}{{\sum_{j=1}^n e^{z_j}}}\) and \(p_1<-\frac{1}{n}\sum_i p_i\).

We have that

\[\frac{\partial p_i}{\partial z_j} = p_i\cdot(\delta_{ij}-p_j),\]

where \(\delta_{ij}\) is the Kronecker delta. The entropy is calculate by \(H(p) = -\sum_{i=1}^n p_i\log(p_i)\).

As the original proposition is equivalent to  \(\frac{\partial H(p)}{\partial z_1} >0\), we have that

\begin{align*}
\frac{\partial H(p)}{\partial z_1} &= \sum_{i=1}^n \frac{\partial H(p)}{\partial p_i}\cdot \frac{\partial p_i}{\partial z_1} \\
        &= \frac{\partial H(p)}{\partial p_1}\cdot \frac{\partial p_1}{\partial z_1} + \sum_{i=2}^n \frac{\partial H(p)}{\partial p_i}\cdot \frac{\partial p_i}{\partial z_1} \\
        &= (-\ln p_1 -1)\cdot p_1\cdot(1-p_1) + \sum_{i=2}^n (-\ln p_i -1)\cdot p_i(-p_1)) \\
        &= p_1\cdot(- \ln p_1 +\sum_{i=1}^n p_i \ln p_i ) \\
        &= p_1\cdot(- \ln p_1 -H(p)). \\
\end{align*}

As the sum of the probabilities of all actions \(p_1+p_2+\cdots+p_n =1\) , we have that

\begin{align*}
  \frac{\partial H(p)}{\partial z_1} &= p_1\cdot(- \ln p_1 -H(p)) \\
  &= p_1\cdot(- (p_1+p_2+\cdots+p_n)\cdot\ln p_1 -H(p)). \\
\end{align*}

The entropy \(H(p)\) will be maximum if and only if \( p_2=p_3=\cdots=p_n=\frac{1-p_1}{n-1}\), which implies that

\begin{align*}
  \frac{\partial H(p)}{\partial z_1}  &= p_1\cdot(- (p_1+p_2+\cdots+p_n)\cdot\ln p_1 -H(p)) \\
  &> p_1\cdot(- (p_1+p_2+\cdots+p_n)\cdot\ln p_1 +p_1\ln p_1 + \frac{1-p_1}{n-1}\cdot\sum_{i=2}^n  \ln \frac{1-p_1}{n-1}) \\
  &= p_1\cdot(- (p_1+(n-1)\cdot \frac{1-p_1}{n-1})\cdot\ln p_1 +p_1\ln p_1 + \frac{1-p_1}{n-1}\cdot(n-1)\cdot \ln \frac{1-p_1}{n-1}) \\
  &= p_1\cdot (1-p_1) \cdot(\ln \frac{1-p_1}{n-1} - \ln p_1) >0.   \\
\end{align*}

\end{proof}

\clearpage
\section{Hyperparameter Experiments}
\label{app-hyper}

\subsection{Hyperparameter for the main training phase}

\begin{figure}[h] 
  \centering
  \begin{subfigure}[b]{0.5\textwidth} 
    \includegraphics[width=\textwidth]{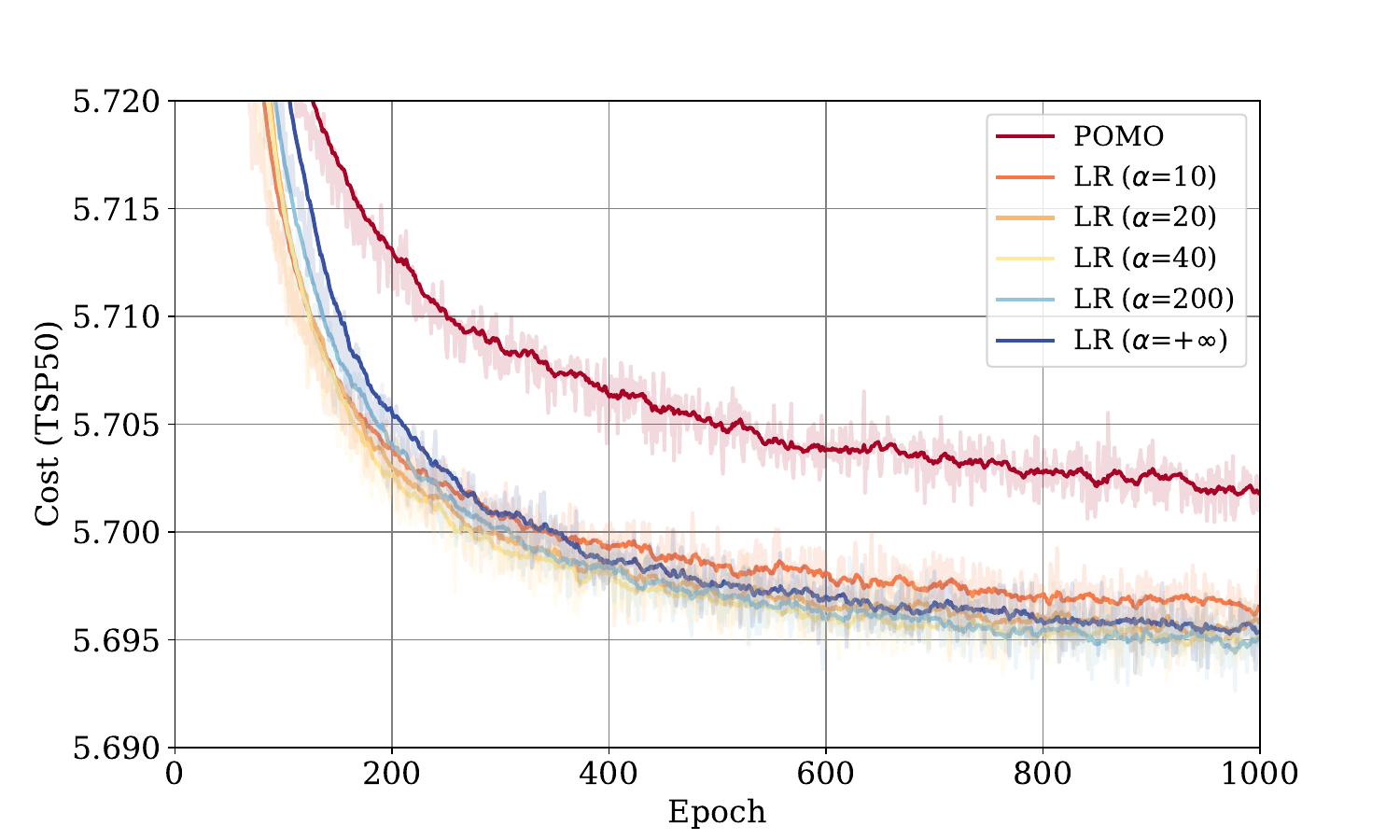}
    \caption{TSP50}
    \label{fig:sub1}
  \end{subfigure}
  \hfill 
  \begin{subfigure}[b]{0.5\textwidth}
    \includegraphics[width=\textwidth]{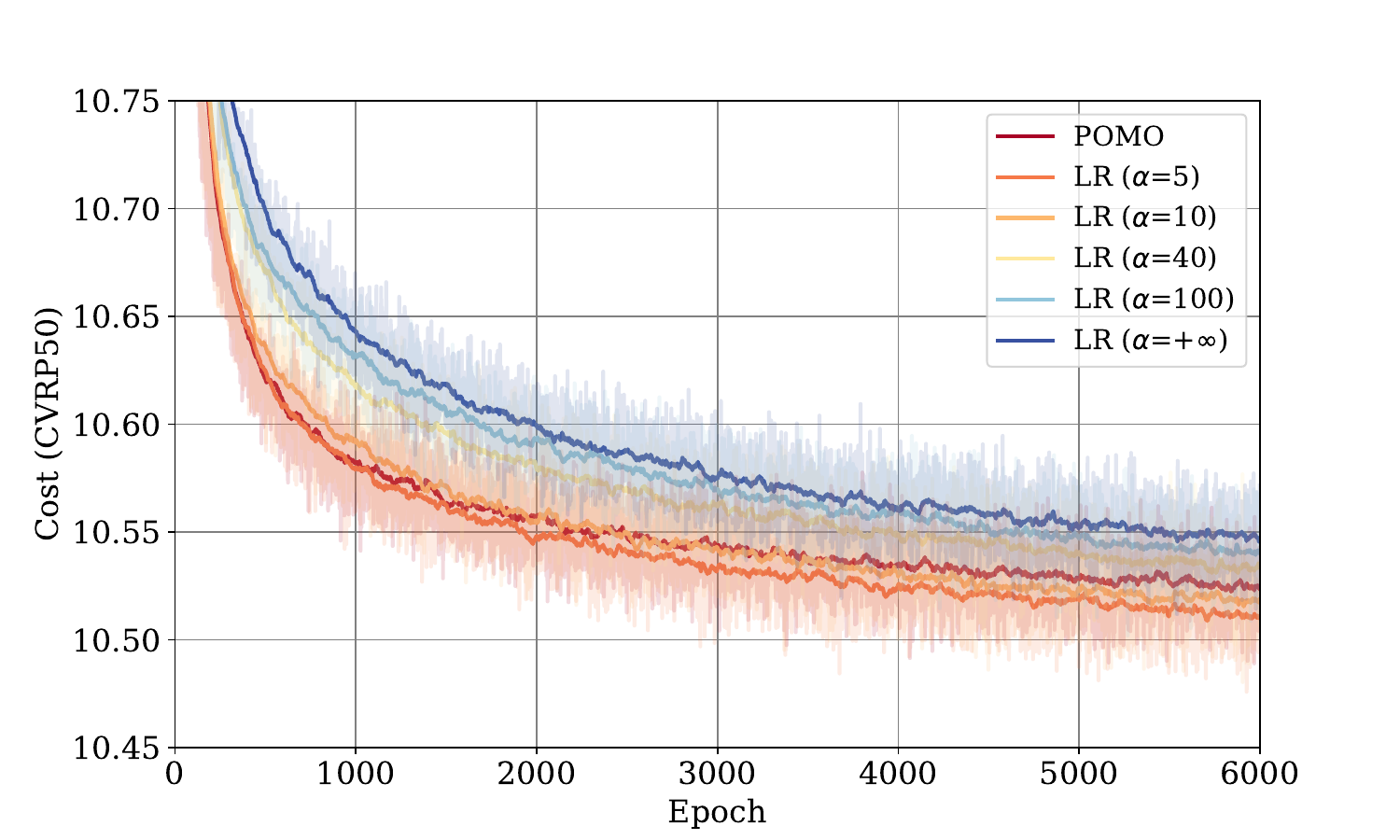}
    \caption{CVRP50}
    \label{fig:sub2}
  \end{subfigure}
  \hfill 
  \begin{subfigure}[b]{0.5\textwidth}
    \includegraphics[width=\textwidth]{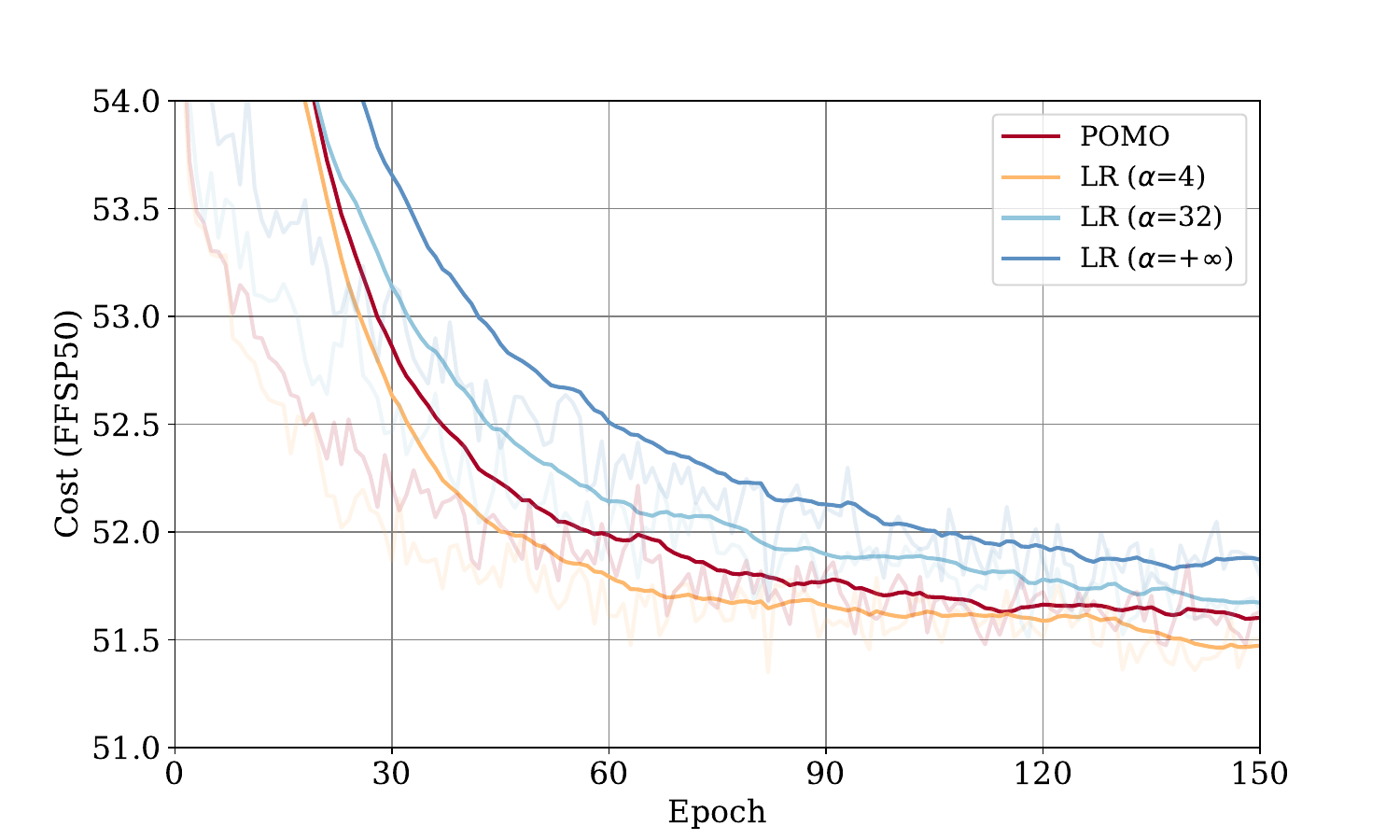}
    \caption{FFSP50}
    \label{fig:sub3}
  \end{subfigure}
  \caption{Comparison of learning curves when choosing different \(\alpha\) for Leader Reward during the main training phase on the TSP, CVRP, and FFSP problem.}
  \label{fig:main3}
\end{figure}

Figure~\ref{fig:main3} shows the learning curves on the TSP, CVRP, and FFSP problem in the case of using different \(\alpha\) as a parameter for Leader Reward in the main training phase. As experiments on problems of size 100 would be too computationally expensive, we give the results for problems of size 50. In TSP, Leader Reward can give a large boost to POMO's performance and is robust to the hyperparameter \(\alpha\). As \(\alpha\) increases, the model gradually converges better, but there is a little negative lift when \(\alpha\) is particularly large. In CVRP and FFSP, using too large an \(\alpha\) will make the model perform worse instead. As we need to balance the exploration and exploitation of the model, a proper \(\alpha\) will make the best use of Leader Reward's performance.

We also notice that Poppy \citep{DBLP:conf/nips/GrinsztajnFSBB23} proposed the policy gradient for populations, which is a little bit like setting \(\alpha = +\infty\) in the Leader Reward when training only one population. However, as shown in the figure, this training method leads to poor model performance. Therefore, we consider poppy and Leader Reward to be two different methods, with the former emphasizing performance improvement through large populations and the latter emphasizing making the model value the leader.



\subsection{Hyperparameter for the final training phase}
\label{hypergamma}

In this section, we test the results of applying Leader Reward to POMO only in the final training phase. Table~\ref{hypergammatable} shows the expected value of the cost, the mean, and the variance of 100 trajectories sampled from the TSP100 problem when choosing different learning rates \(\gamma\). The results show that our method has higher robustness to \(\gamma\), as the costs drop dramatically (from 0.32\% to about 0.11\%) as long as \(\gamma\) is within a reasonable range. 

It's worth noting that our method can significantly increase the variance of solutions (from about 0.05 to about 0.3) and ultimately improve the performance of the model, at the cost of reducing the average quality of the solution. Therefore, controlling \(\gamma\) within a reasonable range can maximize the performance of this method.

\begin{table}[h]
\caption{Hyperparameter experiments of \(\gamma\) in the final training phase}
\label{hypergammatable}
\centering
\begin{tabular}{l||cc|c|c}
\toprule
\multicolumn{1}{c||}{Method}     & Cost(no aug.)    &  Gap          &   Mean \((\mu)\)        &    Variance  \((\sigma)\)   \\
\midrule
POMO        & 7.7891   & 0.316\%   & 7.837      &   0.048         \\
LR \((\gamma=1\times10^{-5})\)        & 7.7739  & 0.121\%     & 7.904      &   0.229         \\
LR \((\gamma=5.5\times10^{-5})\)  & \textbf{7.7726}  & \textbf{0.104\%}    & 7.938    & 0.314         \\
LR \((\gamma=1\times10^{-4})\)  & 7.7728   &  0.106\%    & 7.955    & 0.349        \\
LR \((\gamma=1.5\times10^{-4})\)   & 7.7733   &  0.113\%    & 7.974     & 0.379         \\
\bottomrule
\end{tabular}
\end{table}

\section{Ablation Study}

In order to demonstrate the effectiveness of Leader Reward in both phases, we further conduct an ablation study. Table~\ref{ablation-table} shows the cost and the gap in TSP, CVRP, and FFSP problems after removing some of the phases. The results show that both phases play a key role in improving the performance of the model.

\begin{table}[h]
\caption{Ablation studies of LR}
\label{ablation-table}
\centering
\begin{tabular}{c||cc|cc|cc}
\toprule
\multirow{2}{*}{Method} & \multicolumn{2}{c|}{TSP100} & \multicolumn{2}{c|}{CVRP100} & \multicolumn{2}{c}{FFSP100} \\
                & Cost        & Gap          & Cost         & Gap          & Cost         & Gap          \\
\midrule
LR        & 7.766       & 0.014\%      & 15.729       & 1.064\%      & 89.207       & 1.270\%      \\
\midrule
w/o main phase  & 7.766       & 0.025\%      & 15.732       & 1.085\%      & 89.294       & 1.369\%      \\
w/o final phase & 7.768       & 0.040\%      & 15.748       & 1.191\%      & 89.406       & 1.496\%      \\
w/o both phase              & 7.776       & 0.144\%      & 15.754       & 1.228\%      & 89.745       & 1.881\%  \\
\bottomrule
\end{tabular}
\end{table}

\newpage
\section{Results on other POMO-based models}
\label{mvmoeomni}


We further implement our method on MVMoE \citep{zhou2024mvmoe} and Omni-VRP \citep{Zhou_Wu_Song_Cao_Zhang_2023}. We follow the default setting of the authors and the only change is applying Leader Reward to the advantage function during training. In both models, we set $\alpha=5$ during the main phase. We also report the performance when the final phase is removed. As shown in Table~\ref{MVMoE-table} and Table~\ref{Omni-VRP-table}, both phases bring significant improvement in performance.

\begin{table}[h]
    \caption{Performance on MVMoE ($N=50$)}
    \label{MVMoE-table}
    \centering
    \begin{tabular}{l||rr|rr|rr}
        \toprule
        \multirow{2}{*}{Problem} & \multicolumn{2}{c}{MVMoE} & \multicolumn{2}{c}{LR w/o final phase} & \multicolumn{2}{c}{LR} \\
        & \multicolumn{1}{c}{Cost} & \multicolumn{1}{c}{Gap} & \multicolumn{1}{c}{Cost} & \multicolumn{1}{c}{Gap} & \multicolumn{1}{c}{Cost} & \multicolumn{1}{c}{Gap} \\
        \midrule
        CVRP    & 10.427   & 0.890\%   & 10.423   & 0.846\%   & \textbf{10.416}  & \textbf{0.785\%}  \\
        OVRP    & 6.657    & 2.418\%   & 6.647    & 2.265\%   & \textbf{6.625}   & \textbf{1.937\%}  \\
        VRPB    & 8.167    & 1.495\%   & 8.159    & 1.395\%   & \textbf{8.149}   & \textbf{1.275\%}  \\
        VRPL    & 10.501   & 0.087\%   & 10.496   & 0.039\%   & \textbf{10.486}  & \textbf{-0.057\%} \\
        VRPTW   & 15.000   & 3.412\%   & 14.985   & 3.293\%   & \textbf{14.955}  & \textbf{3.088\%}  \\
        OVRPTW  & 8.964    & 3.217\%   & 8.941    & 2.949\%   & \textbf{8.916}   & \textbf{2.662\%}  \\
        OVRPB   & 6.111    & 6.333\%   & 6.093    & 6.016\%   & \textbf{6.065}   & \textbf{5.536\%}  \\
        OVRPL   & 6.653    & 2.502\%   & 6.641    & 2.323\%   & \textbf{6.624}   & \textbf{2.053\%}  \\
        VRPBL   & 8.175    & 1.808\%   & 8.162    & 1.655\%   & \textbf{8.151}   & \textbf{1.513\%}  \\
        VRPBTW  & 16.019   & 8.589\%   & 15.986   & 8.364\%   & \textbf{15.952}  & \textbf{8.137\%}  \\
        VRPLTW  & 14.927   & 2.362\%   & 14.913   & 2.253\%   & \textbf{14.888}  & \textbf{2.071\%}  \\
        OVRPBL  & 6.098    & 6.215\%   & 6.077    & 5.856\%   & \textbf{6.054}   & \textbf{5.445\%}  \\
        OVRPBTW & 9.490    & 9.350\%   & 9.472    & 9.139\%   & \textbf{9.449}   & \textbf{8.877\%}  \\
        OVRPLTW & 8.965    & 3.385\%   & 8.942    & 3.125\%   & \textbf{8.916}   & \textbf{2.824\%}  \\
        VRPBLTW & 15.941   & 8.748\%   & 15.932   & 8.686\%   & \textbf{15.897}  & \textbf{8.448\%}  \\
        OVRPBTW & 9.514    & 9.644\%   & 9.484    & 9.298\%   & \textbf{9.457}   & \textbf{8.976\%}  \\
        \bottomrule
    \end{tabular}
\end{table}


\begin{table}[h]
    \caption{Performance on Omni-VRP (TSP, zero-shot)}
    \label{Omni-VRP-table}
    \centering
    \begin{tabular}{l||rr|rr|rr}
        \toprule
        \multirow{2}{*}{(Size, Distribution)} & \multicolumn{2}{c}{Omni-VRP} & \multicolumn{2}{c}{LR w/o final phase} & \multicolumn{2}{c}{LR}\\
        & \multicolumn{1}{c}{Cost} & \multicolumn{1}{c}{Gap} & \multicolumn{1}{c}{Cost} & \multicolumn{1}{c}{Gap} & \multicolumn{1}{c}{Cost} & \multicolumn{1}{c}{Gap} \\ 
        \midrule
        $(200, GM_2^5)$     & 9.027  & 2.773\%    & 8.991   & 2.353\%    & \textbf{8.965} & \textbf{2.068\%}   \\
        $(200, R)$          & 8.372  & 2.124\%    & 8.341   & 1.756\%    & \textbf{8.329} & \textbf{1.610\%}   \\
        $(200, E)$          & 8.240  & 1.845\%    & 8.216   & 1.546\%    & \textbf{8.215} & \textbf{1.528\%}   \\
        $(300, U)$          & 13.400 & 3.442\%    & 13.342  & 2.995\%    & \textbf{13.329} & \textbf{2.892\%}  \\
        $(300, GM_3^{10})$  & 9.845  & 3.941\%    & 9.797   & 3.439\%    & \textbf{9.784} & \textbf{3.307\%}   \\
        $(300, GM_7^{50})$  & 5.787  & 2.714\%    & 5.770   & 2.420\%    & \textbf{5.761} & \textbf{2.257\%}   \\
        $(300, R)$          & 10.162 & 3.774\%    & 10.114  & 3.287\%    & \textbf{10.101}     & \textbf{3.149\%}   \\
        $(300, E)$          & 9.801  & 3.384\%    & \textbf{9.764} & \textbf{2.989\%}  & 9.765   & 3.007\%   \\
        $(500, R)$          & 13.428 & 8.366\%    & 13.356  & 7.788\%    & \textbf{13.315}     & \textbf{7.459\%}   \\
        $(500, E)$          & 12.672 & 7.996\%    & 12.616  & 7.519\%    & \textbf{12.589}     & \textbf{7.286\%}   \\
        $(1000, R)$         & 20.374 & 19.214\%   & 20.348  & 19.072\%   & \textbf{20.101}     & \textbf{17.625\%}  \\
        $(1000, E)$         & 18.575 & 18.629\%   & 18.567  & 18.584\%   & \textbf{18.397}     & \textbf{17.502\%}  \\
        \bottomrule
    \end{tabular}
\end{table}

\clearpage
\newpage

\section{Results on TSPLib and CVRPLib}
\label{tsplib}

This section gives a performance evaluation of our method on TSPLib \citep{reinelt1991tsplib} and CVRPLib \citep{uchoa2017new}, where we choose instances with size $N<250$. Both POMO and our model are pre-trained on $N=100$ and evaluated in Table~\ref{tsp-table} and Table~\ref{cvrp-table}. In this experiment, we report results under greedy strategy and instance augmentation. As shown below, our method outperforms POMO on both datasets. See Table~\ref{TSPLib-table} and Table~\ref{CVRPLib-table}.

\begin{table}[h]
    \caption{Performance comparison on TSPLib instances.}
    \label{TSPLib-table}
    \centering
    \begin{tabular}{l||r|rr|rr}
        \toprule
        \multirow{2}{*}{Instance} & \multirow{2}{*}{Optimal} & \multicolumn{2}{c}{POMO}           & \multicolumn{2}{c}{LR+POMO}          \\
                &        & Cost            & Gap              & Cost            & Gap             \\
        \midrule
        berlin52& 7542   & 7545            & 0.04\%           & \textbf{7544}   & \textbf{0.03\%} \\
        bier127 & 118282 & 128660          & 8.77\%           & \textbf{120606} & \textbf{1.96\%} \\
        ch130   & 6110   & 6120            & 0.16\%           & \textbf{6117}   & \textbf{0.11\%} \\
        ch150   & 6528   & 6562            & 0.52\%           & \textbf{6559}   & \textbf{0.47\%} \\
        d198    & 15780  & \textbf{18508}  & \textbf{17.29\%} & 18967           & 20.20\%         \\
        eil101  & 629    & 641             & 1.91\%           & \textbf{640}    & \textbf{1.75\%} \\
        eil51   & 426    & 430             & 0.94\%           & \textbf{429}    & \textbf{0.70\%} \\
        eil76   & 538    & \textbf{544}    & \textbf{1.12\%}  & \textbf{544}    & \textbf{1.12\%} \\
        kroA100 & 21282  & 21370           & 0.41\%           & \textbf{21285}  & \textbf{0.01\%} \\
        kroA150 & 26524  & \textbf{26709}  & \textbf{0.70\%}  & 26783           & 0.98\%          \\
        kroA200 & 29368  & \textbf{29831}  & \textbf{1.58\%}  & 30088           & 2.45\%          \\
        kroB100 & 22141  & 22212           & 0.32\%           & \textbf{22210}  & \textbf{0.31\%} \\
        kroB150 & 26130  & 26435           & 1.17\%           & \textbf{26331}  & \textbf{0.77\%} \\
        kroB200 & 29437  & \textbf{29876}  & \textbf{1.49\%}  & 30004           & 1.93\%          \\
        kroC100 & 20749  & 20787           & 0.18\%           & \textbf{20760}  & \textbf{0.05\%} \\
        kroD100 & 21294  & 21473           & 0.84\%           & \textbf{21412}  & \textbf{0.55\%} \\
        kroE100 & 22068  & 22167           & 0.45\%           & \textbf{22137}  & \textbf{0.31\%} \\
        lin105  & 14379  & 14454           & 0.52\%           & \textbf{14419}  & \textbf{0.28\%} \\
        pr107   & 44303  & \textbf{44585}  & \textbf{0.64\%}  & 44663           & 0.81\%          \\
        pr124   & 59030  & 59246           & 0.37\%           & \textbf{59075}  & \textbf{0.08\%} \\
        pr136   & 96772  & \textbf{97521}  & \textbf{0.77\%}  & 97562           & 0.82\%          \\
        pr144   & 58537  & \textbf{58802}  & \textbf{0.45\%}  & 59403           & 1.48\%          \\
        pr152   & 73682  & \textbf{74596}  & \textbf{1.24\%}  & 75218           & 2.08\%          \\
        pr226   & 80369  & \textbf{83281}  & \textbf{3.62\%}  & 87103           & 8.38\%          \\
        pr76    & 108159 & \textbf{108159} & \textbf{0.00\%}  & \textbf{108159} & \textbf{0.00\%} \\
        rat195  & 2323   & 2512            & 8.14\%           & \textbf{2432}   & \textbf{4.69\%} \\
        rat99   & 1211   & 1234            & 1.90\%           & \textbf{1224}   & \textbf{1.07\%} \\
        rd100   & 7910   & \textbf{7910}   & \textbf{0.00\%}  & \textbf{7910}   & \textbf{0.00\%} \\
        st70    & 675    & \textbf{677}    & \textbf{0.30\%}  & \textbf{677}    & \textbf{0.30\%} \\
        ts225   & 126643 & 132623          & 4.72\%           & \textbf{128711} & \textbf{1.63\%} \\
        tsp225  & 3861   & 4111            & 6.48\%           & \textbf{4093}   & \textbf{6.01\%} \\
        u159    & 42080  & 42480           & 0.95\%           & \textbf{42460}  & \textbf{0.90\%} \\
        \midrule
        Avg Gap &        &                 & 2.12\%           & \textbf{}       & \textbf{1.95\%} \\
        \bottomrule
    \end{tabular}
\end{table}

\begin{table}
    \caption{Performance comparison on CVRPLib instances.}
    \label{CVRPLib-table}
    \centering
    \begin{tabular}{l||r|rr|rr}
        \toprule
        \multirow{2}{*}{Instance} & \multirow{2}{*}{Optimal} & \multicolumn{2}{c}{POMO}           & \multicolumn{2}{c}{LR+POMO}          \\
                &        & Cost            & Gap              & Cost           & Gap              \\
        \midrule
        X-n247-k50 & 37274  & \textbf{42420}  & \textbf{13.81\%} & 44217          & 18.63\%          \\
        X-n242-k48 & 82751  & \textbf{89185}  & \textbf{7.78\%}  & 89669          & 8.36\%           \\
        X-n237-k14 & 27042  & 31272           & 15.64\%          & \textbf{29363} & \textbf{8.58\%}  \\
        X-n233-k16 & 19230  & 21119           & 9.82\%           & \textbf{20979} & \textbf{9.10\%}  \\
        X-n228-k23 & 25742  & 29354           & 14.03\%          & \textbf{28758} & \textbf{11.72\%} \\
        X-n223-k34 & 40437  & \textbf{43804}  & \textbf{8.33\%}  & 43877          & 8.51\%           \\
        X-n219-k73 & 117595 & \textbf{122208} & \textbf{3.92\%}  & 124756         & 6.09\%           \\
        X-n214-k11 & 10856  & 11763           & 8.35\%           & \textbf{11741} & \textbf{8.15\%}  \\
        X-n209-k16 & 30656  & \textbf{32399}  & \textbf{5.69\%}  & 32410          & 5.72\%           \\
        X-n204-k19 & 19565  & \textbf{20976}  & \textbf{7.21\%}  & 21284          & 8.79\%           \\
        X-n200-k36 & 58578  & \textbf{62102}  & \textbf{6.02\%}  & 62272          & 6.31\%           \\
        X-n195-k51 & 44225  & 50303           & 13.74\%          & \textbf{49632} & \textbf{12.23\%} \\
        X-n190-k8  & 16980  & \textbf{18067}  & \textbf{6.40\%}  & 18389          & 8.30\%           \\
        X-n186-k15 & 24145  & \textbf{25742}  & \textbf{6.61\%}  & 25709          & 6.48\%           \\
        X-n181-k23 & 25569  & 26978           & 5.51\%           & \textbf{26606} & \textbf{4.06\%}  \\
        X-n176-k26 & 47812  & 52883           & 10.61\%          & \textbf{52644} & \textbf{10.11\%} \\
        X-n172-k51 & 45607  & \textbf{50356}  & \textbf{10.41\%} & 51538          & 13.00\%          \\
        X-n167-k10 & 20557  & \textbf{21297}  & \textbf{3.60\%}  & 21649          & 5.31\%           \\
        X-n162-k11 & 14138  & 14985           & 5.99\%           & \textbf{14969} & \textbf{5.88\%}  \\
        X-n157-k13 & 16876  & 18302           & 8.45\%           & \textbf{17692} & \textbf{4.84\%}  \\
        X-n153-k22 & 21220  & 24356           & 14.78\%          & \textbf{24050} & \textbf{13.34\%} \\
        X-n148-k46 & 43448  & \textbf{47621}  & \textbf{9.60\%}  & 47861          & 10.16\%          \\
        X-n143-k7  & 15700  & \textbf{16380}  & \textbf{4.33\%}  & 16533          & 5.31\%           \\
        X-n139-k10 & 13590  & 14080           & 3.61\%           & \textbf{13906} & \textbf{2.33\%}  \\
        X-n134-k13 & 10916  & \textbf{11315}  & \textbf{3.66\%}  & 11334          & 3.83\%           \\
        X-n129-k18 & 28940  & 29569           & 2.17\%           & \textbf{29373} & \textbf{1.50\%}  \\
        X-n125-k30 & 55539  & \textbf{58421}  & \textbf{5.19\%}  & 60122          & 8.25\%           \\
        X-n120-k6  & 13332  & 14570           & 9.29\%           & \textbf{13882} & \textbf{4.13\%}  \\
        X-n115-k10 & 12747  & 13878           & 8.87\%           & \textbf{13351} & \textbf{4.74\%}  \\
        X-n110-k13 & 14971  & 15160           & 1.26\%           & \textbf{15149} & \textbf{1.19\%}  \\
        X-n106-k14 & 26362  & \textbf{26967}  & \textbf{2.29\%}  & 27423          & 4.02\%           \\
        X-n101-k25 & 27591  & \textbf{29288}  & \textbf{6.15\%}  & 29878          & 8.29\%           \\
        \midrule
        Avg Gap &        &  & 7.60\%           & \textbf{}      & \textbf{7.41\%}  \\
        \bottomrule
    \end{tabular}
\end{table}




\clearpage
\newpage
\section*{NeurIPS Paper Checklist}

\begin{enumerate}

\item {\bf Claims}
    \item[] Question: Do the main claims made in the abstract and introduction accurately reflect the paper's contributions and scope?
    \item[] Answer: \answerYes{} 
    \item[] Justification: They do.

\item {\bf Limitations}
    \item[] Question: Does the paper discuss the limitations of the work performed by the authors?
    \item[] Answer: \answerYes{} 
    \item[] Justification:  There is a limitation on \(\alpha\) and \(\gamma\), and we discussed them in Section~\ref{methods} and Appendix~\ref{app-hyper}.

\item {\bf Theory Assumptions and Proofs}
    \item[] Question: For each theoretical result, does the paper provide the full set of assumptions and a complete (and correct) proof?
    \item[] Answer: \answerYes{} 
    \item[] Justification: We provide proof in Appendix~\ref{app_proof}.

    \item {\bf Experimental Result Reproducibility}
    \item[] Question: Does the paper fully disclose all the information needed to reproduce the main experimental results of the paper to the extent that it affects the main claims and/or conclusions of the paper (regardless of whether the code and data are provided or not)?
    \item[] Answer: \answerYes{} 
    \item[] Justification: We disclose it in Section~\ref{experiments}.

\item {\bf Open access to data and code}
    \item[] Question: Does the paper provide open access to the data and code, with sufficient instructions to faithfully reproduce the main experimental results, as described in supplemental material?
    \item[] Answer: \answerYes{} 
    \item[] Justification:  We provide the code in the supplementary material.

\item {\bf Experimental Setting/Details}
    \item[] Question: Does the paper specify all the training and test details (e.g., data splits, hyperparameters, how they were chosen, type of optimizer, etc.) necessary to understand the results?
    \item[] Answer: \answerYes{} 
    \item[] Justification: We specify them in Section~\ref{experiments}.

\item {\bf Experiment Statistical Significance}
    \item[] Question: Does the paper report error bars suitably and correctly defined or other appropriate information about the statistical significance of the experiments?
    \item[] Answer: \answerNo{} 
    \item[] Justification: The main experiment is based on a large number of inferences, generated from a fixed seed for consistency with previous work.

\item {\bf Experiments Compute Resources}
    \item[] Question: For each experiment, does the paper provide sufficient information on the computer resources (type of compute workers, memory, time of execution) needed to reproduce the experiments?
    \item[] Answer: \answerYes{} 
    \item[] Justification: We provide it in Section~\ref{experiments}.
    
\item {\bf Code Of Ethics}
    \item[] Question: Does the research conducted in the paper conform, in every respect, with the NeurIPS Code of Ethics \url{https://neurips.cc/public/EthicsGuidelines}?
    \item[] Answer: \answerYes{} 
    \item[] Justification: It does.

\item {\bf Broader Impacts}
    \item[] Question: Does the paper discuss both potential positive societal impacts and negative societal impacts of the work performed?
    \item[] Answer: \answerYes{} 
    \item[] Justification: We discuss them in Section~\ref{conclusion}.
    
\item {\bf Safeguards}
    \item[] Question: Does the paper describe safeguards that have been put in place for responsible release of data or models that have a high risk for misuse (e.g., pretrained language models, image generators, or scraped datasets)?
    \item[] Answer: \answerNA{} 
    \item[] Justification: The paper poses no such risks.

\item {\bf Licenses for existing assets}
    \item[] Question: Are the creators or original owners of assets (e.g., code, data, models), used in the paper, properly credited and are the license and terms of use explicitly mentioned and properly respected?
    \item[] Answer: \answerYes{} 
    \item[] Justification: We credit them and mention the license.

\item {\bf New Assets}
    \item[] Question: Are new assets introduced in the paper well documented and is the documentation provided alongside the assets?
    \item[] Answer: \answerNA{} 
    \item[] Justification: We do not release new assets.

\item {\bf Crowdsourcing and Research with Human Subjects}
    \item[] Question: For crowdsourcing experiments and research with human subjects, does the paper include the full text of instructions given to participants and screenshots, if applicable, as well as details about compensation (if any)? 
    \item[] Answer: \answerNA{} 
    \item[] Justification: The paper does not involve crowdsourcing nor research with human subjects.

\item {\bf Institutional Review Board (IRB) Approvals or Equivalent for Research with Human Subjects}
    \item[] Question: Does the paper describe potential risks incurred by study participants, whether such risks were disclosed to the subjects, and whether Institutional Review Board (IRB) approvals (or an equivalent approval/review based on the requirements of your country or institution) were obtained?
    \item[] Answer: \answerNA{} 
    \item[] Justification: The paper does not involve crowdsourcing nor research with human subjects.

\end{enumerate}

\end{document}